\documentclass[letterpaper]{article} % DO NOT CHANGE THIS
\usepackage{aaai24}  % DO NOT CHANGE THIS
\usepackage{times}  % DO NOT CHANGE THIS
\usepackage{helvet}  % DO NOT CHANGE THIS
\usepackage{courier}  % DO NOT CHANGE THIS
\usepackage[hyphens]{url}  % DO NOT CHANGE THIS
\usepackage{graphicx} % DO NOT CHANGE THIS
\usepackage{natbib}  % DO NOT CHANGE THIS AND DO NOT ADD ANY OPTIONS TO IT
\usepackage{caption} % DO NOT CHANGE THIS AND DO NOT ADD ANY OPTIONS TO IT
\usepackage{algorithm}
\usepackage{algorithmic}

\usepackage{makecell}
\usepackage{multirow}
\usepackage{threeparttable}
\usepackage{amsmath} % for equation*
\usepackage{booktabs} % for toprule, midrule, bottomrule

\usepackage{newfloat}
\usepackage{listings}
\DeclareCaptionStyle{ruled}{labelfont=normalfont,labelsep=colon,strut=off} % DO NOT CHANGE THIS
\floatstyle{ruled}
\newfloat{listing}{tb}{lst}{}
\floatname{listing}{Listing}
\title{CFEVER: A Chinese Fact Extraction and VERification Dataset}
\author{
    Ying-Jia Lin, Chun-Yi Lin, Chia-Jen Yeh, Yi-Ting Li, \\
    Yun-Yu Hu, Chih-Hao Hsu, Mei-Feng Lee, Hung-Yu Kao
}
\affiliations{
    Department of Computer Science and Information Engineering, National Cheng Kung University, Tainan, Taiwan\\
    yingjia.lin.public@gmail.com, hykao@mail.ncku.edu.tw
}

\usepackage{bibentry}
\begin{document}

\maketitle

\begin{abstract}
    We present CFEVER, a \textbf{C}hinese dataset designed for \textbf{F}act \textbf{E}xtraction and \textbf{VER}ification.
    CFEVER comprises 30,012 manually created claims based on content in Chinese Wikipedia.
    Each claim in CFEVER is labeled as ``Supports", ``Refutes", or ``Not Enough Info" to depict its degree of factualness.
    Similar to the FEVER dataset, claims in the ``Supports" and ``Refutes" categories are also annotated with corresponding evidence sentences sourced from single or multiple pages in Chinese Wikipedia.
    Our labeled dataset holds a Fleiss' kappa value of 0.7934 for five-way inter-annotator agreement.
    In addition, through the experiments with the state-of-the-art approaches developed on the FEVER dataset and a simple baseline for CFEVER, we demonstrate that our dataset is a new rigorous benchmark for factual extraction and verification, which can be further used for developing automated systems to alleviate human fact-checking efforts.
    CFEVER is available at \url{https://ikmlab.github.io/CFEVER}.
\end{abstract}

\section{Introduction}

Fact verification involves assessing the truthfulness of claims presented in text or speech.
In recent years, the popularization of media platforms has accelerated the spread of misinformation, making fact verification a critical task to prevent the public from being exposed to false information.
However, the process of fact verification typically involves extensive searches and assessments from many potential sources conducted by journalists, which is time-consuming and labor-intensive \cite{guo-etal-2022-survey}.
Consequently, it is imperative to develop automated fact verification systems to speed up the verification process.

In recent years, researchers have been building fact-verification systems with deep neural networks \cite{weibo16}, where the models are provided with a claim and required to determine whether the claim is true or false.
% introduce FEVER
A well-established dataset for fact verification is FEVER (Fact Extraction and VERification) \cite{thorne-etal-2018-fever} and its shared task \cite{thorne-etal-2018-fact}.
FEVER asks models to extract factual sentences as evidence from the fixed Wikipedia database and provide a verdict for a given claim.
Therefore, the task serves for both verification and evidence extraction for claims.
% issue
Though FEVER has been widely used as a benchmark for fact verification, the dataset and most of the other fact verification datasets \cite{wang-2017-liar,hanselowski-etal-2019-richly, schuster-etal-2021-get} are created in English.
% The lack of datasets hinders the development of fact-verification systems in other languages.
The dissemination of rumors and fake news is a serious problem in East Asia, especially in China.
Compared to English, text written in Chinese tends to be more ambiguous and nuanced, making it more difficult for people to identify misinformation.
Therefore, it is critical to create a clean and high-quality dataset as supervised data for Chinese fact verification.
To achieve this goal, \citet{hu-etal-2022-chef} proposed the CHEF dataset from the sources of the fact-checking and news websites.
However, compared to English fact verification datasets \cite{augenstein-etal-2019-multifc,kotonya-toni-2020-explainable-automated},
the size of CHEF is relatively small, making training and evaluating fact verification systems challenging.
A potential alternative for building Chinese fact verification systems is to take advantage of the multilingual fact-checking dataset \cite{MuMiN}.
Nevertheless, the Chinese claims in the dataset are still limited in size.

\begin{table}[t]
    \centering
    \includegraphics[width=1.0\columnwidth]{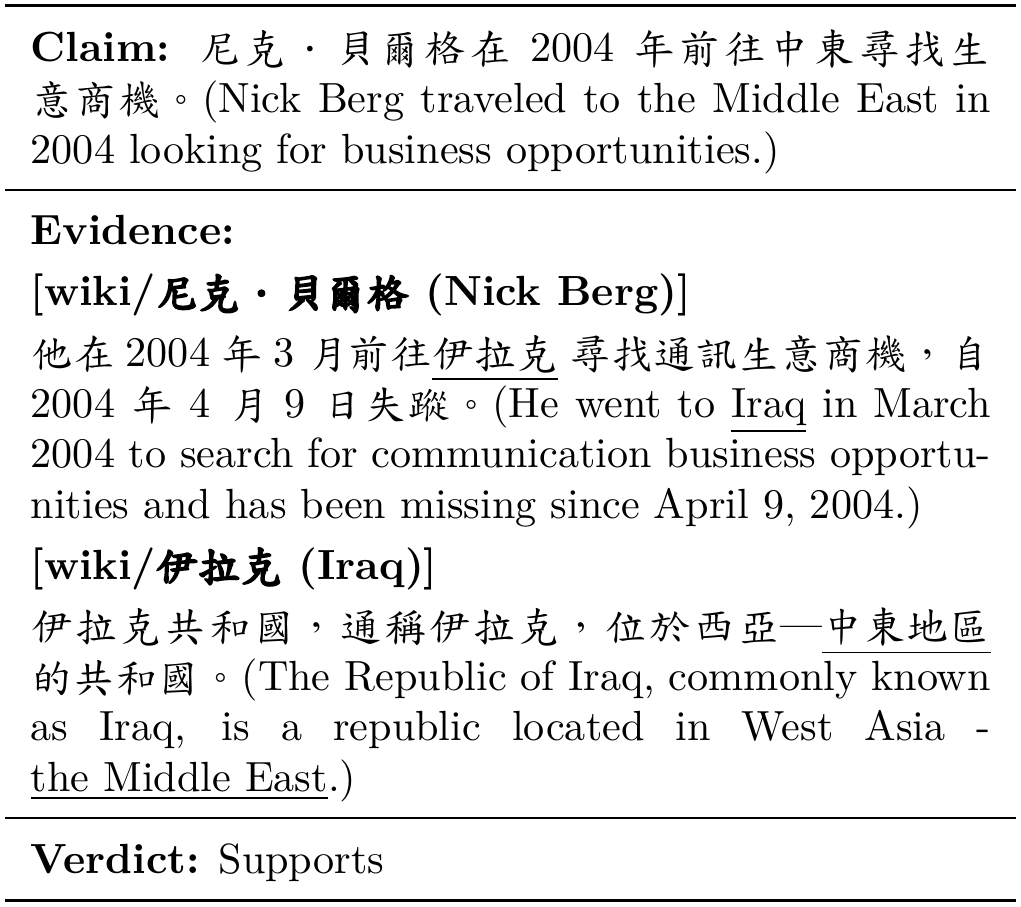}
    \caption{An example from the CFEVER dataset. Underlined words indicate knowledge that should be verified from other pages in Wikipedia.}
    \label{tab:example}
\end{table}

\begin{table*}[t]
    \centering
    \resizebox{1.99\columnwidth}{!}{
        \begin{tabular}{lcccccc}
            \toprule
            \multirow{3}{*}{Dataset}      & \multirow{3}{*}{\#Claims}          & \multirow{3}{*}{Source}  & \multirow{3}{*}{Language}             & \multicolumn{3}{c}{Task}                            \\
            \cmidrule{5-7}
                                          &                                    &                          &                                       & Document                 & Evidence  & Claim        \\
                                          &                                    &                          &                                       & Retrieval                & Retrieval & Verification \\
            \midrule
            Weibo-16 \cite{weibo16}       & 4,664                              & Weibo                    & Chinese                               & No                       & No        & Yes*         \\
            Weibo-20 \cite{weibo20}       & 6,362                              & Weibo                    & Chinese                               & No                       & No        & Yes*         \\
            MuMiN \cite{MuMiN}            & 1,283\textsuperscript{\textdagger} & Tweets                   & Multi\textsuperscript{\textdaggerdbl} & No                       & No        & Yes          \\
            CHEF \cite{hu-etal-2022-chef} & 10,000                             & \makecell{Fact-checking/                                                                                               \\News websites} & Chinese & No & Yes & Yes \\
            \midrule
            CFEVER                        & 30,012                             & Wikipedia                & Chinese                               & Yes                      & Yes       & Yes          \\
            \bottomrule
        \end{tabular}}
    \footnotesize
    % * Rumor detection task.

    * Rumor detection task. \textsuperscript{\textdagger} Including Chinese and other languages. \textsuperscript{\textdaggerdbl} Multi-lingual dataset.\\

    \caption{Comparison of CFEVER with other fact verification datasets in Chinese.}
    \label{tab:dataset}
\end{table*}

In this work, we present CFEVER, a \textbf{C}hinese dataset for \textbf{F}actual \textbf{E}xtraction and \textbf{VER}ification.
Following FEVER's construction process \cite{thorne-etal-2018-fever}, we first built a fact database with the fixed version of the Chinese Wikipedia dump.
Then, we hired several workers from our university to alter the extracted sentences from the Wikipedia pages and label each claim with ``Supports", ``Refutes", or ``Not Enough Info" based on the Wikipedia pages in the fact database.
The relevant factual sentences from the fact database were also annotated by our workers as evidence for the claims belonging to the first two categories.
Our CFEVER dataset includes 30,012 claims in total, which are three times the size of CHEF \cite{hu-etal-2022-chef}.
To evaluate our labeling quality, we report the five-way inter-annotator agreement of 0.7934 in Fleiss $\kappa$ \cite{fleiss1971measuring} using 6.45\% claims of our dataset, while the scores in FEVER and CHEF are 0.6841 and 0.74, measured from 4\% and 3\% of claims from the datasets, respectively.

To thoroughly assess the challenges of CFEVER, following the FEVER task \cite{thorne-etal-2018-fact}, we conduct experiments on the three stages: document retrieval, sentence retrieval, and claim verification.
We test the performance for each stage and the full-pipeline setting with the state-of-the-art approaches \cite{stammbach-2021-evidence,dehaven-scott-2023-bevers} developed on FEVER and our simple baseline \cite{bm25,BERTfever} along with oracle experiments to dive into the challenges of CFEVER.
Further extensive analysis even reveals the characteristics and difficulty of CFEVER.
In summary, we list our contributions as follows:

\begin{itemize}
    \item We present CFEVER, the currently largest Chinese dataset for evidence-based fact verification.
    \item The five-way inter-annotator agreement of CFEVER in Fleiss $\kappa$ indicates that the dataset was built with high label consistency among claims.
    \item Extensive experiments on CFEVER show that the proposed dataset can serve as a challenging benchmark for future research or development on Chinese fact verification.
\end{itemize}

\section{Related Work}
\subsection{FEVER}
\citet{thorne-etal-2018-fact} created the task of Fact Extraction and VERification (FEVER) along with the dataset of the same name \cite{thorne-etal-2018-fever}.
FEVER is the currently largest fact verification dataset that contains 185,445 claims.
Each of the claims was first sampled from the sentences in the introductory sections of approximately 50,000 popular pages and then revised by human annotators.
In another round of annotation, the annotators label the claims as ``Supports", ``Refutes", or ``Not Enough Information", and discover evidence sentences from Wikipedia pages.
The following year, \citet{thorne-etal-2019-fever2} introduced the FEVER 2.0 task to first generate adversarial claims to fool the existing verification systems built with the FEVER dataset \cite{thorne-etal-2018-fever}.
Then the task required participants to improve the systems to prevent such adversarial attacks.
More recently, \citet{aly2021feverous} proposed FEVEROUS and extended the fact verification task for verifying claims with the information in structured data, such as tables in Wikipedia.

\citet{danfever} follow the annotation process of FEVER and propose a new FEVER dataset \cite{thorne-etal-2018-fever} in Danish.
\citet{jiang-etal-2020-hover} report that 87\% of the claims in the FEVER dataset \cite{thorne-etal-2018-fever} require only a single Wikipedia page for verification, which does not support real-world situations where misinformation may come from multiple articles.
Thus, Jiang et al. (2020) propose a new dataset containing 26K claims for multi-hop reasoning based on the building process of FEVER.

\subsection{Chinese Fact Verification}
We compare CFEVER with other Chinese fact verification datasets in Table \ref{tab:dataset}.
Existing Chinese fact verification datasets mainly focus on rumor detection, such as the Weibo-16 \cite{weibo16} and Weibo-20 datasets \cite{weibo20}.
In this work, we focus on both the fact extraction and verification tasks, which are different from the binary claim detection in rumor detection \cite{guo-etal-2022-survey}.
The dataset closest to our work is CHEF \cite{hu-etal-2022-chef}, which is a pilot Chinese dataset for evidence-based fact-checking.
There are two main differences between CHEF and our dataset.
First, CHEF is created from the sources of fact-checking websites, while our dataset is created from Wikipedia.
Second, there is no document retrieval process in the task of CHEF, with candidate evidence sentences provided for each claim in the dataset.
In contrast, we follow FEVER \cite{thorne-etal-2018-fever} to provide a fixed fact database and ask models to first extract relevant documents from Wikipedia before verifying the claims with the evidence sentences.
% TODO: fix `along with`
We consider our approach, along with FEVER \cite{thorne-etal-2018-fever}, to be more realistic for real-world fact verification scenarios.

\section{Dataset Construction}
We followed the labeling approach of FEVER \cite{thorne-etal-2018-fever} to create the CFEVER dataset and adapted the annotation platform\footnote{\url{https://github.com/awslabs/fever/tree/master/fever-annotations-platform}} publicly released by \citet{thorne-etal-2018-fever} for our construction task.
The annotation process consists of two stages: \textbf{claim generation} and \textbf{claim annotation}.
Both stages are conducted by human workers recruited from our university.
The two stages of annotation are conducted separately.
To distinguish the two stages, we refer to the workers in the claim generation stage as \textbf{writers} and the ones in the claim annotation stage as \textbf{annotators}, based on the characteristics of their tasks.
Before the construction process, we first describe our method to prepare the Wikipedia data and the fact database.

\subsection{Preparation for the Wikipedia Data}
We used the December 2022 dump of the Chinese Wikipedia and extracted the text from the introductory section from each page following the pre-processing method of FEVER \cite{thorne-etal-2018-fever}.
The raw Wikipedia comprises articles in both Traditional and Simplified Chinese.
To unify the data, we processed the text with the open-source software OpenCC\footnote{\url{https://github.com/BYVoid/OpenCC}} to convert the text to Traditional Chinese.
Then, the processed data containing 1,187,751 pages were fixed to serve as the fact database for CFEVER.
For the next stage of data construction, we created a source page pool based on the processed Wikipedia pages.
The source page pool includes the 500 most visited Chinese Wikipedia pages\footnote{\url{https://pageviews.wmcloud.org/topviews/?project=zh.wikipedia.org}} worldwide in 2022, 10,000 Taiwanese pages, and 3,000 random pages.
All the claims in our dataset were created based on the pages in the source page pool.

\subsection{Claim Generation}
At this stage, writers are responsible for writing claims based on the Wikipedia pages in the source page pool.
At each time, each writer was given one extracted sentence randomly sampled from the introductory section of a page in the source page pool.
In addition, Wikipedia pages related to the given sentence based on the hyperlinks in the raw Wiki data were also provided to the writer.
Then, we asked a writer to first generate a TRUE claim based on the given extracted sentence and the information in the relevant pages without any learned human knowledge.
The provided relevant pages were used to help writers come up with diverse claims, which may also result in complex claims that require multi-step reasoning from different pages in the verification task.
After that, the writer was asked to generate six variants of the TRUE claim:
\begin{itemize}
    \item \textbf{Rephrasing}: A TRUE claim should be rephrased to a different sentence with the same meaning.
    \item \textbf{Negation}: A TRUE claim should be negated without simple negation words, such as ``not''.
    \item \textbf{Entity substitution at a similar level}: An entity in a TRUE claim should be substituted with another one similar to the original entity.
    \item \textbf{Entity substitution at a disjointed level}: An entity in a TRUE claim should be substituted with another one disjointed to the original entity.
    \item \textbf{Specification}: A TRUE claim should be narrowed down with more specific concepts.
    \item \textbf{Generalization}: A TRUE claim should be generalized with more abstract concepts.
\end{itemize}

\begin{figure}[t]
    \centering
    \includegraphics[width=0.99\columnwidth]{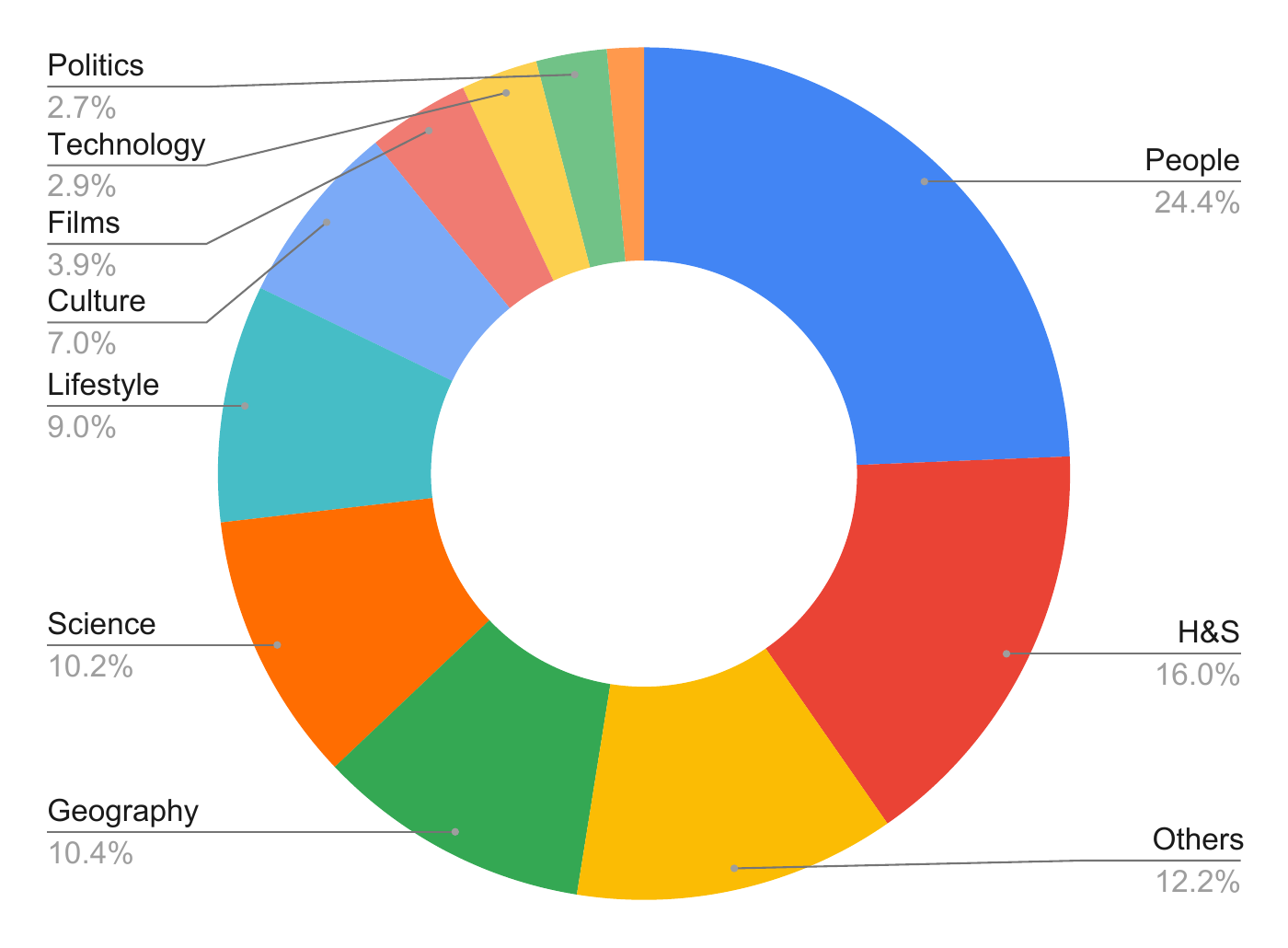} % Reduce the figure size so that it is slightly narrower than the column. Don't use precise values for figure width.This setup will avoid overfull boxes.
    \caption{Domain distribution of the generated claims in our dataset. H\&S refers to the domain of Humanities \& Social Sciences.}
    \label{fig1}
\end{figure}

During claim generation, the writers were asked not to generate claims with their own learned knowledge.
They should write claims solely based on the information in the given extracted sentence and the relevant pages.
The reason behind this step is to help generate verifiable claims and maintain the quality of the generated claims among different writers.
% domains
We also measured the domains of the claims in CFEVER.
To prevent misclassification based on the domains of source Wiki pages, we asked the writers to select a category from the pre-defined domains for each claim they generated.
The final domain distribution of the generated claims is shown in Figure \ref{fig1}.

% We encouraged the writers to generate claims that can be referenced by multiple pages for increasing the complexity of the claims.

\subsection{Claim Annotation}
Once a claim was generated, an annotator was asked to label the claim as ``Supports", ``Refutes", or ``Not Enough Info".
For the first two categories, the annotator must also find the sentences as evidence from the fact database.
To achieve such a process, at each time of annotation, four kinds of materials are provided to the annotator in default:
\begin{itemize}
    \item \textbf{Claim}: The claim generated by the writer in the claim generation stage.
    \item \textbf{Page name}: The title of the original page from which the claim was generated.
    \item \textbf{Original sentences}: The sentences in the introductory section of the original page which the claim was generated from.
    \item \textbf{Relevant pages}: The pages related to the original page based on the hyperlinks in the raw Wiki data.
\end{itemize}
% we provide the sentences in the introductory section of the original page which the claim was generated from for the annotator to verify the factualness of the given claim.
% The relevant pages to the original page of the claim in the claim generation step were also available to the annotator.
If none of the sentences provided in default can be selected as evidence for the claim, an annotator can also search the fact database with a Wiki page name as a keyword.
After passing the keyword to the annotation platform, the sentences in the introductory section of the Wikipedia page will show up at the annotation interface to be selectable as evidence by an annotator.
Note that a claim may become ``Supports" or ``Refutes" based on multiple sentences.
We encouraged the annotators to find as many sentences as possible to support or refute a claim from the fact database.
Once the annotator considered no factual sentences in our fact database could support or refute a claim, the claim was labeled as ``Not Enough Info."

Some claims were generated based on the original extracted sentence along with the content from the relevant pages in the claim generation stage.
These claims require evidence referenced from multiple pages.
For the example in Table \ref{tab:example}, the term ``Middle East" in the claim cannot be inferred directly from the sentence mentioning ``Iraq."
Another sentence in the ``Iraq" page should be selected as a part of the evidence, even though the relationship between the two terms is common knowledge.
Therefore, for these claims, we asked annotators to combine two or more sentences from different pages as complete evidence to support or refute the claim using the annotation platform.

\subsection{Workers}
We recruited nine writers and annotators from our university in total.
All of the workers were native Chinese speakers.
Among them, three workers were from the College of Liberal Arts, with two of them focused on the claim generation task only, and the remaining six were from the College of Engineering.
The workers were trained by the authors for the annotation tasks and guidelines until they were able to undergo the annotation process correctly and independently.
%TODO: unify the annotation terms

\begin{table}[t]
    \centering
    \resizebox{.99\columnwidth}{!}{
        \begin{tabular}{c|c|c|c}
            \toprule
            Split             & Training                    & Development                 & Test                        \\
            \midrule
            \midrule
            Total Claims      & 24,012                      & 3,000                       & 3,000                       \\
            % \hline
            Num of SUP        & 11,085                      & 1,000                       & 1,000                       \\
            Num of REF        & 7,113                       & 1,000                       & 1,000                       \\
            Num of NEI        & 5,814                       & 1,000                       & 1,000                       \\
            % \hline
            % Avg. Claim length (words) & 18.6 & 19.0 & 19.0 \\
            Avg. Claim Length & 33.19                       & 34.04                       & 34.04                       \\
            \midrule
            Avg. Evidence     & \multirow{2}{*}{1.57 sents} & \multirow{2}{*}{1.53 sents} & \multirow{2}{*}{1.55 sents} \\
            per Claim         &                             &                             &                             \\
            \midrule
            Avg. Evidence     & 1.13                        & 1.14                        & 1.12                        \\
            Pages per Claim   & (88.41\%)                   & (87.35\%)                   & (89.45\%)                   \\
            \bottomrule
        \end{tabular}}
    \caption{Dataset statistics of CFEVER in different splits. SUP indicates the ``Supports" class, REF stands for ``Refutes", and NEI represents ``Not Enough Information".
        The Avg. Claim Length is the average number of characters in a claim.
        The Avg. Evidence per Claim is the average number of evidence sentences for a claim in the ``Supports" or ``Refutes" class, and the ratio in the parentheses is the proportion of claims with evidence from one single page.
    }
    \label{table3}
\end{table}

\subsection{Data Validation}
To evaluate the consistency of the class labels among the annotators, we randomly sampled 1,936 claims (6.45\% of our total dataset) and asked five annotators from our workers to label them.
The five-way inter-annotator agreement over the 1,936 claims shows a score of 0.7934 in Fleiss $\kappa$ \cite{fleiss1971measuring}, which is higher than the score of 0.6841 measured with 7,506 claims (4\% from total) in FEVER \cite{thorne-etal-2018-fever} and the one of 0.74 with 310 claims (3\% from total) in the CHEF dataset \cite{hu-etal-2022-chef}.
% TODO
To evaluate the correctness of the evidence sentences, we randomly sampled another 700 claims (2.33\% of the dataset) for being reviewed by the authors.
We discovered that 84.4\% of the claims are annotated correctly with correct evidence sentences.

\subsection{Dataset Statistics}
There are 30,012 claims in the CFEVER dataset.
We split 80\%, 10\%, and 10\% of the claims into the training, development, and test sets, respectively.
The statistics of the dataset are shown in Table \ref{table3}.
The number of claims in the three categories is balanced in the development and test sets to ensure that the performance of the models is not biased towards any category during evaluations.
% To make data distributions among the three sets more consistent,
The average claim length (character level) and the average number of evidence sentences per claim are also similar among the three splits.
We also report the ratio of claims with evidence from a single page in Table \ref{table3}, with 88.41\%, 87.35\%, and 89.45\% of the claims whose evidence can be found in a single page in the three sets.
These ratios are close to the statistic of FEVER \cite{thorne-etal-2018-fever} reported by \citet{jiang-etal-2020-hover}, where 87\% of the claims require one page for verification.

\section{Baseline Systems}
Following \citet{thorne-etal-2018-fact}, our task requires a model to retrieve evidence from the Wikipedia fact database and perform verification for each claim.
This section introduces the approaches we test for CFEVER in three stages: document retrieval, sentence retrieval, and recognizing textual entailment (RTE) for claim verification.
The systems we build involves two full-pipeline methods\footnote{BERT (\url{https://huggingface.co/hfl/chinese-bert-wwm-ext}) is used for BEVERS and our baseline unless otherwise noted.} for the three stages: one simple baseline proposed by ourselves and the state-of-the-art approach \cite{dehaven-scott-2023-bevers} developed for FEVER \cite{thorne-etal-2018-fever}.
We also test CFEVER with the sentence retrieval approach proposed by \citet{stammbach-2021-evidence}.
We provide essential details in this section, and more information is available on the CFEVER website.

\subsection{Our Baseline}
% For the FEVER task \cite{thorne-etal-2018-fact}, numerous approaches have been proposed to improve the performance of claim verification \cite{BERTfever,KGAT,proofver,dehaven-scott-2023-bevers} or even the retrieval tasks \cite{hanselowski-etal-2018-ukp,BERTfever,stammbach-2021-evidence,dehaven-scott-2023-bevers}.
To understand the difficulty and behaviors of CFEVER, we first design a baseline with simple components to explore the dataset.

\subsubsection{Document Retrieval}
For evidence-based claim verification, relevant pages should be discovered for each claim to extract evidence.
Many studies for the FEVER task \cite{thorne-etal-2018-fact} adopt TF-IDF \cite{thorne-etal-2018-fever} or the search with the MediaWiki API \cite{hanselowski-etal-2018-ukp} for document retrieval.
Following \citet{jiang-etal-2020-hover}, we use BM25 \cite{bm25} for retrieving relevant pages for each claim.
Our implementation was based on Elasticsearch\footnote{\url{https://github.com/elastic/elasticsearch}}, and the representations were built with the Wikipedia pages from our fact database.

\subsubsection{Sentence Retrieval}
% \subsubsection{Token Retriever \cite{stammbach-2021-evidence}}
% Instead of classifying each sentence in a Wikipedia page as evidence or non-evidence, \citet{stammbach-2021-evidence} treats sentence retrieval as a token-level classification problem,
% where a model has to predict 1 for each token inside evidence sentences and 0 for the tokens belonging to non-evidence sentences.
% Such an approach requires a model to process long sequences from input claim-article pairs.
% Thus, \citet{stammbach-2021-evidence} adopts BigBird \cite{NEURIPS2020_c8512d14} as the encoder for sentence retrieval.

% \subsubsection{Pointwise Retriever}
After retrieving relevant pages for each claim, we perform sentence retrieval to select evidence sentences from the pages.
Inspired by \citet{hanselowski-etal-2018-ukp} and \citet{BERTfever}, we implement a pointwise approach for sentence retrieval with BERT \cite{devlin-etal-2019-bert} to classify each claim-sentence pair in binary.
Positive pairs are created using claims and their corresponding gold evidence sentences.
In contrast, negative pairs consist of claims paired with non-evidence sentences, which are sampled from the predicted pages acquired during the document retrieval phase.

\subsubsection{Recognizing Textual Entailment}
We verify each claim with the evidence sentences by recognizing textual entailment (RTE).
Following \citet{hanselowski-etal-2018-ukp,nie2019combining}, we first concatenate a claim with its top five evidence sentences
% \begin{equation*}\small
%     \textrm{[CLS] Claim\space [SEP]\space Evidence1\space [SEP]\space ...\space Evidence5\space [SEP]}
% \end{equation*}
% , where each evidence sentence is separated by the [SEP] token.
and then fine-tune the BERT model \cite{devlin-etal-2019-bert} for the three-class RTE task.

\subsection{BEVERS}
\subsubsection{Document Retrieval}
The second approach is based on the BEVERS \cite{dehaven-scott-2023-bevers}, which is the state-of-the-art full-pipeline system for FEVER \cite{thorne-etal-2018-fact}.
BEVERS uses a hybrid approach to include both the search results from Wikisearch \cite{hanselowski-etal-2018-ukp} and the TF-IDF method \cite{thorne-etal-2018-fever}.
Such a hybrid approach was also adopted by \citet{stammbach-2021-evidence}.
% Such the hybrid approach was also adopted by \citet{jiang-etal-2021-exploring-listwise} with BM25 in substitution of TF-IDF.
Note that BEVERS replaced the MediaWiki API in the approach of \citet{hanselowski-etal-2018-ukp} with a fuzzy string search system.

\subsubsection{Sentence Retrieval}
Besides the document retrieval approach, \citet{dehaven-scott-2023-bevers} also proposed a competitive sentence retrieval approach.
BEVERS extends the binary pointwise approach \cite{hanselowski-etal-2018-ukp} with an additional ternary classification task for classifying a claim-sentence pair into ``Supports", ``Refutes", or ``Not Enough Info", as an initial set of predicted evidence sentences.
Then, BEVERS used the results from the initial set to explore more evidence sentences from the hyperlinks in a Wikipedia article, which was called ``re-retrieval."
Finally, all extracted sentences from these two steps are ranked to yield the final evidence sentences for each claim.

\subsubsection{Recognizing Textual Entailment}
In addition to the concatenation-based approach \cite{hanselowski-etal-2018-ukp,nie2019combining} for performing claim verification with concatenated evidence, the singleton-based approach \cite{malon-2018-team,BERTfever} was also proposed to classify each claim-evidence pair individually.
In this setting, each claim will have multiple scores for each evidence sentence.
Then, the scores are aggregated based on the rules \cite{malon-2018-team, BERTfever} to obtain the final prediction.
BEVERS adopts a mixture of both approaches \cite{dehaven-scott-2023-bevers}.
They first fine-tune DeBERTa-V2-XL \cite{he2021deberta} pre-trained on MNLI \cite{mnli} for each of the approaches and train an additional gradient boosting classifier \cite{friedman2001greedy} for aggregating the final predictions.

\subsection{Stammbach}
\citet{stammbach-2021-evidence} treats sentence retrieval as a token-level classification problem,
where a model must predict 1 for each token within evidence sentences and 0 for the tokens belonging to non-evidence sentences.
Such an approach requires a model to process long sequences from input claim-article pairs.
Thus, \citet{stammbach-2021-evidence} adopts BigBird \cite{NEURIPS2020_c8512d14} as the encoder.
Since this approach was proposed for sentence retrieval, we only test this approach for retrieving evidence with the ground-truth documents.

\section{Evaluation Metrics}
For document retrieval and sentence retrieval, we report the performance in recall (\%).
Our recall evaluation metric is designed to assess the model's ability to correctly predict at least one complete set of evidence pages during document retrieval and, similarly, at least one complete set of evidence sentences during sentence retrieval, for each data instance.
For claim verification in RTE, following \citet{thorne-etal-2018-fact}, we report performance in accuracy (\%) and FEVER Score (\%).
The latter is a strict measure of accuracy, requiring a model to correctly predict at least one complete evidence set for each claim.
For implementing the evaluation metrics, we use the script from \citet{dehaven-scott-2023-bevers} for document retrieval.
For sentence retrieval and RTE, we utilize the official scoring tool from \citet{thorne-etal-2018-fact}.

\section{Results and Analysis}
\subsection{Results in Different Stages}
To thoroughly analyze CFEVER, we list the results for the different stages in Table \ref{table:main}.
% The other metrics, such as precision and F1 scores for document and sentence retrieval stages, are provided in Appendix.
For document retrieval, we observe that BEVERS achieves more than 90\% of recall,  demonstrating that the approach is able to find correct articles for most of the given claims.
The performance difference between our simple baseline (BM25) and BEVERS is primarily due to the hybrid approach employed in BEVERS, which combines the search predictions from the TF-IDF method and the fuzzy string search.
% However, BM25 remains a competitive baseline for document retrieval.
% The precision and F1 scores for document retrieval can be found in Appendix.

For sentence retrieval, there’s a huge performance gap between our simple baseline and BEVERS.
This may result from the training with an additional ternary classification task and the employment of the re-retrieval technique in BEVERS, whose score is also far ahead of the other baselines close to our sentence selection method on FEVER.
However, BEVERS scored 94.41\% in recall reported in their paper \cite{dehaven-scott-2023-bevers} for FEVER, showing that our dataset remains a challenge for BEVERS.
% The precision and F1 scores at this stage can also be found in Appendix.

For claim verification and the full-pipeline setting, BEVERS obtains only 69.73\% for the label accuracy, which is much lower than their score (80.24\%) reported for the FEVER dataset \cite{dehaven-scott-2023-bevers}.
Since the classification is based on the evidence extracted in the sentence retrieval stage, the scores for claim verification will be affected if a model cannot identify correct evidence sentences.
Still, BEVERS significantly outperforms our simple baseline by 8.56\% in label accuracy and 12.33\% in FEVER Score.

We also test the performance of GPT-3.5 \cite{gpt3_5} and GPT-4 \cite{openai2023gpt}\footnote{The prompts we used can be found from the CFEVER website.} for claim verification with the test set using the \textit{zero-shot} and few-shot settings.
For the few-shot setting, we sample three labeled claims from the training set with the same domain as the input claim for each class.
From Table \ref{table:main}, we find that the claims in CFEVER are challenging for both models, and the performance can be slightly improved with the few-shot setting.

\subsection{Oracle Results}
To analyze the difficulty of CFEVER further, we also report the oracle results for the last two stages in Table \ref{table:gold}.
For the oracle setting in sentence retrieval, the gold documents of each claim are provided for the models.
As for the oracle setting in claim verification, the models take gold evidence sentences for each claim as inputs.
The results show that both methods achieve more than 95\% in recall for the oracle setting in sentence retrieval.
However, there are still 15\% and 10\% error rates in claim verification for our baseline and BEVERS ($\textrm{RoBERTa}_\textrm{Large}$\footnote{\url{https://huggingface.co/hfl/chinese-roberta-wwm-ext-large}}), showing that certain claims in our dataset pose verification challenges.
We notice that Stammbach performs worse than the other two baselines for sentence retrieval.
The results may be affected by the Chinese BigBird we used\footnote{\url{https://huggingface.co/Lowin/chinese-bigbird-base-4096}}, since there is no official Chinese version of BigBird \cite{NEURIPS2020_c8512d14}.

\begin{table}[t]
    \centering
    \begin{tabular}{l|lc}
        \toprule
        Task (metric)                   & System                      & Score (\%) \\
        \midrule\midrule
        Doc retrieval                   & Our baseline                & 87.65      \\
        (Recall)                        & BEVERS$^a$                  & 92.60      \\
        \midrule
        Sent retrieval                  & Our baseline                & 76.65      \\
        (Recall)                        & BEVERS$^a$                  & 86.60      \\
        \midrule
        \multirow{6}{*}{RTE (Accuracy)} & Our baseline                & 61.17      \\
                                        & BEVERS$^a$                  & 69.73      \\
                                        & GPT-3.5 (\textit{zeroshot}) & 43.17      \\
                                        & GPT-3.5 (\textit{3-shot})   & 44.20      \\
                                        & GPT-4 (\textit{zeroshot})   & 47.23      \\
                                        & GPT-4 (\textit{3-shot})     & 48.40      \\
        \midrule
        Full pipeline                   & Our baseline                & 52.47      \\
        (FEVER Score)                   & BEVERS$^a$                  & 64.80      \\
        \bottomrule
    \end{tabular}
    \caption{Results on the different stages. $a$: \citet{dehaven-scott-2023-bevers}.}
    \label{table:main}
\end{table}

\begin{table}[t]
    \centering
    \begin{tabular}{l|lc}
        \toprule
        Task (metric) & System                                           & Score (\%) \\
        \midrule\midrule
        \multirow{3}{*}{\makecell{Sent retrieval                                      \\(Recall)}}            & Our baseline                                     & 95.90      \\
                      & BEVERS$^a$                                       & 95.20      \\
                      & Stammbach$^b$                                    & 83.55      \\
        \midrule
        \multirow{3}{*}{\makecell{RTE                                                 \\(Accuracy)}}            & Our baseline     & 85.10      \\
                      & BEVERS$^{a}$ ($\textrm{BERT}_\textrm{Base}$)     & 88.50      \\
                      & BEVERS$^{a}$ ($\textrm{RoBERTa}_\textrm{Large}$) & 90.33      \\
        \bottomrule
    \end{tabular}
    \caption{Oracle results on sentence retrieval and the task of recognizing textual entailment (RTE). $a$: \citet{dehaven-scott-2023-bevers}. $b$: \citet{stammbach-2021-evidence}.}
    \label{table:gold}
\end{table}

\subsection{Analysis for Claims with Evidence from Multiple Pages}
After testing the two methods in different stages, we further analyze the performance on claims with evidence labeled from different numbers of Wikipedia pages for the full-pipeline setting in accuracy and FEVER Score.
The results in Table \ref{table:pages} show that the performance of both methods degrades significantly for the claims with evidence from more pages.
For the claims with more than three pages of evidence, the two methods achieve unsatisfactory performance in FEVER Score.
In summary, we identify that about 10\% claims with evidence from multiple pages in our dataset are challenging to verify.
Even BEVERS can only get about 70\% of the FEVER Score for the claims with evidence from single pages.

\begin{table}[t]
    \centering
    \resizebox{.997\columnwidth}{!}{
        \begin{tabular}{l|ccc}
            \toprule
            \multirow{2}{*}{Methods} & \multicolumn{3}{c}{\#Pages (ratio)}                                     \\
                                     & 1 (89.45\%)                         & 2 (9.35\%)    & $\geq$ 3 (1.20\%) \\
            \midrule
            Ours                     & 67.75 / 57.97                       & 55.08 / 15.51 & 50.00 / 0.00      \\
            BEVERS$^a$               & 73.95 / 70.54                       & 63.10 / 21.93 & 58.33 / 16.67     \\
            \bottomrule
        \end{tabular}}
    \caption{
        Full-pipeline results in accuracy / FEVER Score (\%) for the claims with evidence from different number of pages. ``Ratio" indicates the proportion of the claims from the test set (not counted for the claims of ``Not Enough Information'') in each group. $a$: \citet{dehaven-scott-2023-bevers}.
    }
    \label{table:pages}
\end{table}

\subsection{Analysis for Claims with Different Numbers of Evidence Sentences}
\begin{table}[t]
    \centering
    \resizebox{.999\columnwidth}{!}{
        \begin{tabular}{l|cccc}
            \toprule
            \multirow{3}{*}{Methods} & \multicolumn{4}{c}{\#Evidence sentences (ratio)}                                           \\
                                     & 1                                                & 2           & 3           & $\geq$ 4    \\
                                     & (61.75\%)                                        & (26.40\%)   & (10.15\%)   & (1.70\%)    \\
            \midrule
            Ours                     & 64.1 / 55.9                                      & 72.2 / 56.3 & 65.5 / 36.9 & 61.8 / 11.8 \\
            BEVERS$^a$               & 70.8 / 67.8                                      & 75.6 / 65.3 & 80.3 / 57.1 & 55.9 / 26.5 \\
            \bottomrule
        \end{tabular}}
    \caption{
        Full-pipeline results in accuracy / FEVER Score (\%) for the claims with different numbers of evidence sentences. ``Ratio" indicates the proportion of the claims from the test set (not counted for the claims of ``Not Enough Information'') in each group. a: \citet{dehaven-scott-2023-bevers}.
    }
    \label{table:sents}
\end{table}

Since each claim in our dataset can have multiple evidence sentences, we also analyze the performance on claims with different numbers of evidence sentences.
The results in Table \ref{table:sents} show that both methods perform better on the claims with two and three evidence sentences in the full-pipeline setting.
This may result from the fact that the claims with one evidence sentence are usually shorter ones, containing less information to be verified.
Additionally, the claims with more than three evidence sentences are usually much longer, requiring more correct evidence sentences to be identified and thus more difficult for the models.

\subsection{Analysis for Claims from Different Domains}
\begin{figure}[t]
    \centering
    \includegraphics[width=1.0\columnwidth]{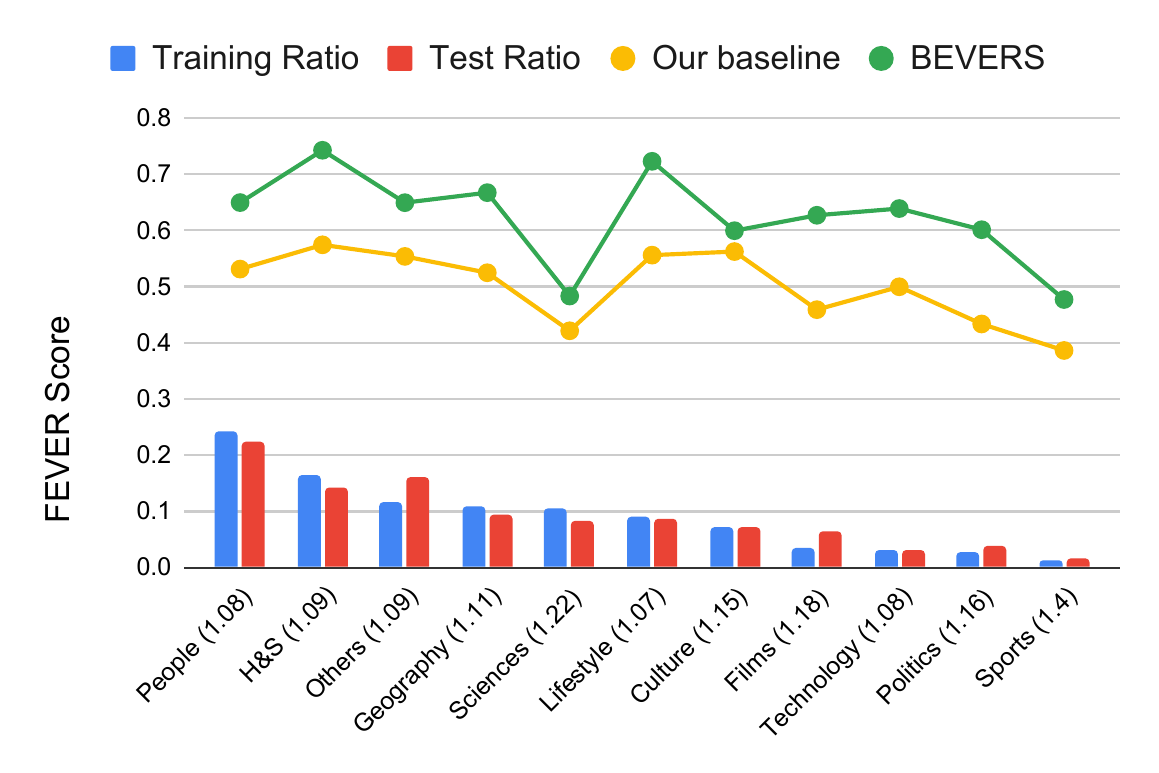} % Reduce the figure size so that it is slightly narrower than the column. Don't use precise values for figure width.This setup will avoid overfull boxes.
    \caption{Performance comparisons for the claims of different domains in the full-pipeline setting. H\&S refers to the domain of Humanities \& Social Sciences. Values in the parentheses are the average number of evidence pages.}
    \label{fig2}
\end{figure}

As the chart in Figure \ref{fig1} shows, the claims in our dataset were labeled into 11 domains by our writers during the claim generation process.
Figure \ref{fig2} shows the performance of the two methods in FEVER Score on claims from different domains.
We discover that both of the methods have lower performance for the domains with fewer training claims, such as ``Sports,'' ``Technology,'' and ``Politics.''
We also observe that the number of Wikipedia pages of the evidence for claims in the ``Sports'' domain is higher than the numbers of all the other domains, which may result in lower performance and match our findings in Table \ref{table:pages}.

\subsection{Analysis for Claims in Different Lengths}
\begin{figure}[t]
    \centering
    \includegraphics[width=1.0\columnwidth]{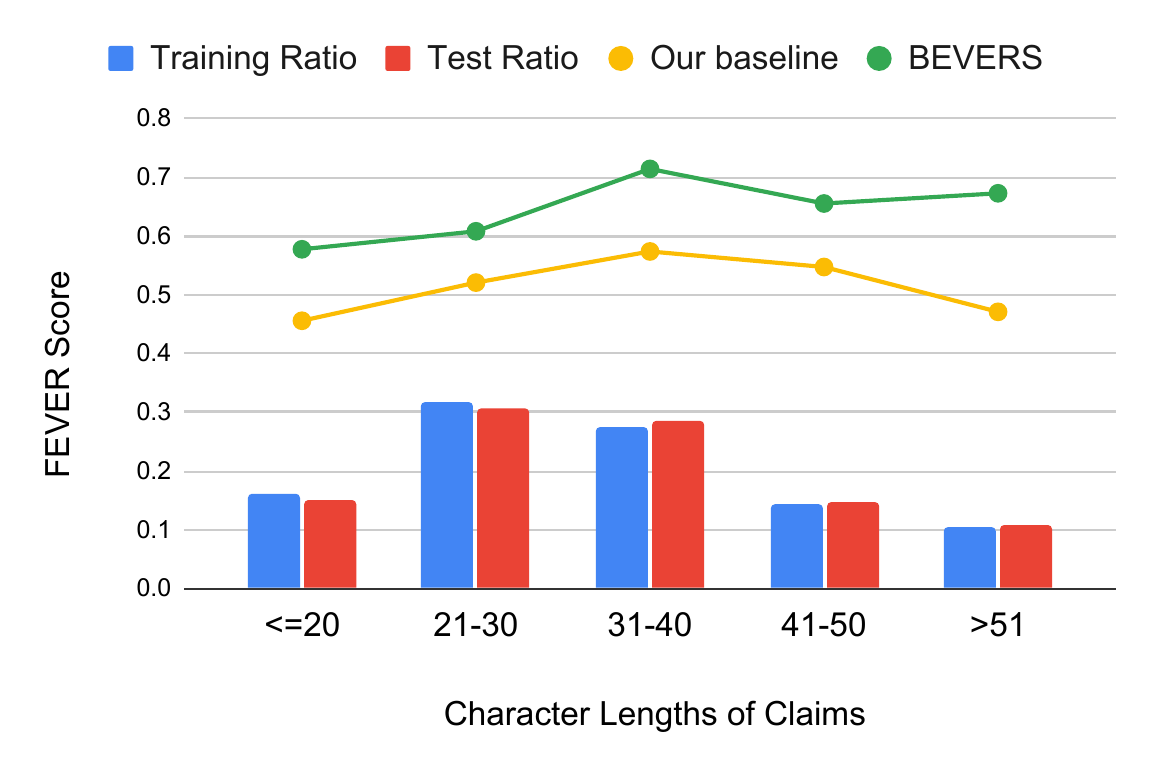} % Reduce the figure size so that it is slightly narrower than the column. Don't use precise values for figure width.This setup will avoid overfull boxes.
    \caption{Performance comparisons with different lengths of claims in the full-pipeline setting. ``Ratio" indicates the proportion for each group of the claims from the test set.}
    \label{fig:length}
\end{figure}

In this section, we analyze the performance of the two methods on claims with different lengths in the full-pipeline setting.
We divide the claims in the test set into five groups according to the number of characters in the claims and show the results in Figure \ref{fig:length}.
We observe that our simple baseline performs worse on the claims longer than 51 characters, while BEVERS remains stable.
Furthermore, the two methods perform better on the medium length claims with 31-40 characters.
These results again indicate that the length of the claims is an important factor in the claim verification task.
% which confirms the results in Table \ref{table:sents}.

\section{Discussion}
We discuss two limitations of our dataset in this section.
First, although CFEVER is currently the largest Chinese dataset for fact extraction and verification, it is still much smaller than the English FEVER dataset.
Data size is an important factor for model performance and generalization.
We hope to increase the data scale in the future while maintaining the high quality of annotation.
Second, CFEVER may not be a perfect dataset for training models to handle complex reasoning tasks.
As we show in Table \ref{tab:dataset}, most of the claims in CFEVER require evidence extracted from only one Wikipedia page.
This issue has also been reported for FEVER \cite{jiang-etal-2020-hover}.
We hope to expand our dataset with more complex claims that require evidence from multiple pages in the future.

\section{Conclusion}
This paper introduces CFEVER, a new Chinese dataset for Fact Extraction and VERification.
Following the FEVER task \cite{thorne-etal-2018-fact}, CFEVER forms a verification task that requires models to verify the claims into ``Supports", ``Refutes", and ``Not Enough Information".
In addition, for the first two categories, models are also required to extract the evidence sentences from our fact database composed of Chinese Wikipedia pages.
We carefully validate the quality of the dataset and obtain an inter-annotator agreement of 0.7934 in Fleiss $\kappa$ for the class label consistency.
Though the experiments with the simple baseline designed by ourselves and the state-of-the-art method \cite{dehaven-scott-2023-bevers} developed on FEVER \cite{thorne-etal-2018-fever}, we believe that CFEVER is a challenging dataset to serve as a benchmark on Chinese claim verification and fact extraction.
% The more than 30K claims with evidence from Wikipedia along with the fact database will be publicly available and can be used for future research or developments of Chinese fact-checking systems.

% \subsubsection{Section Headings.}
% Sections should be arranged and headed as follows:
% \begin{enumerate}
%     \item Main content sections
%     \item Appendices (optional)
%     \item Ethical Statement (optional, unnumbered)
%     \item Acknowledgements (optional, unnumbered)
%     \item References (unnumbered)
% \end{enumerate}

\section*{Acknowledgements}
This work was supported by the National Science and Technology Council of Taiwan, under Grant NSTC 112-2223-E-006-009.
We thank the anonymous reviewers for their insightful comments.
We also extend our deepest gratitude to all of our data annotators for their hard work and dedication during the data construction process.

% \bigskip
% \noindent Thank you for reading these instructions carefully. We look forward to receiving your electronic files!

\bibliography{aaai24}
\clearpage

\appendix
\section{Appendix}
\subsection{Implementation Details}
For our baseline, we retrieved the top ten documents for each claim using BM25 in document retrieval; we retrieved the top five sentences as evidence for each claim in sentence retrieval.
For the former, we mainly chose the setting with the highest recall score, which we provide a comparison in Table \ref{table:bm25}.
For the latter, we followed the statistics of CFEVER (Table \ref{table:sents}) to include potential evidence for all claims with the top five sentences.
To implement our baseline in the sentence retrieval and RTE stages, we use HuggingFace Transformers\footnote{\url{https://huggingface.co/docs/transformers/}} and PyTorch\footnote{\url{https://pytorch.org}} (version: 1.13.1+cu117) to build the models.
The hyperparameter sets are present in Table \ref{table:params} for our baseline.
The final scores were determined based on the performance of the development set using grid searches.

\begin{table}[htbp]
    \centering
    \resizebox{0.8\columnwidth}{!}{
        \begin{tabular}{c|ccc}
            \toprule
            Top-$k$ & Recall         & Precision & F1 Score \\
            \midrule
            1       & 69.05          & 75.25     & 72.02    \\
            3       & 80.10          & 29.67     & 43.30    \\
            5       & 83.85          & 18.73     & 30.62    \\
            7       & 86.00          & 13.74     & 23.69    \\
            10      & \textbf{87.65} & 9.79      & 17.62    \\
            \bottomrule
        \end{tabular}}
    \caption{Document retrieval performance on the test set using BM25 with different top-$k$ values.}
    \label{table:bm25}
\end{table}

\begin{table}[htbp]
    \centering
    \resizebox{1.0\columnwidth}{!}{
        \begin{tabular}{llc}
            \toprule
            Stage                           & Hyperparameter   & Value                \\
            \midrule
            \multirow{5}{*}{Sent Retrieval} & learning rate    & \{2e-5, 3e-5, 5e-5\} \\
                                            & num of epochs    & \{1, 2\}             \\
                                            & batch size       & 64                   \\
                                            & negative samples & 50\%                 \\
                                            & warmup ratio     & 10\% training steps  \\
            \midrule
            \multirow{4}{*}{RTE}            & learning rate    & \{3e-5, 5e-5, 7e-5\} \\
                                            & num of epochs    & \{2, 3\}             \\
                                            & batch size       & 32                   \\
                                            & warmup ratio     & 10\% training steps  \\
            \bottomrule
        \end{tabular}}
    \caption{Hyperparameters we used for our baseline.}
    \label{table:params}
\end{table}
For implementing BEVERS, we followed the instructions and ran their code available on GitHub\footnote{\url{https://github.com/mitchelldehaven/bevers}}.
To make fair comparisons between the two baseline systems, we used the same pre-trained BERT model\footnote{\url{https://huggingface.co/hfl/chinese-bert-wwm-ext}} in the sentence retrieval and RTE stages for BEVERS and our baseline.
For Stammbach \cite{stammbach-2021-evidence}, we used the official code\footnote{\url{https://github.com/dominiksinsaarland/document-level-FEVER}} with the Chinese BigBird\footnote{\url{https://huggingface.co/Lowin/chinese-bigbird-base-4096}} for the sentence retrieval stage.

\subsection{Details for ChatGPT}
We used the GPT-3.5 Turbo and the GPT-4 Turbo models\footnote{https://platform.openai.com/docs/models/} with the following versions for the experiments in Table \ref{table:main}:
\begin{itemize}
    \item GPT-3.5: \textit{gpt-3.5-turbo}
    \item GPT-4: \textit{gpt-4-1106-preview}
\end{itemize}
The prompts we used for GPT-3.5 and GPT-4 are shown in Table \ref{tab:prompt1} and Table \ref{tab:prompt2}.
\noindent Note that we asked GPT-4 to use Chinese Wikipedia for verification in the prompt, since we found that GPT-4 performed better with the requirement (underlined in Table \ref{tab:prompt2}).
Also note that though we provide the translation of the prompts in Table \ref{tab:prompt1} and Table \ref{tab:prompt2}, the English part was not included in the experiments.

\begin{table}[htbp]
    \centering
    \includegraphics[width=1.0\columnwidth]{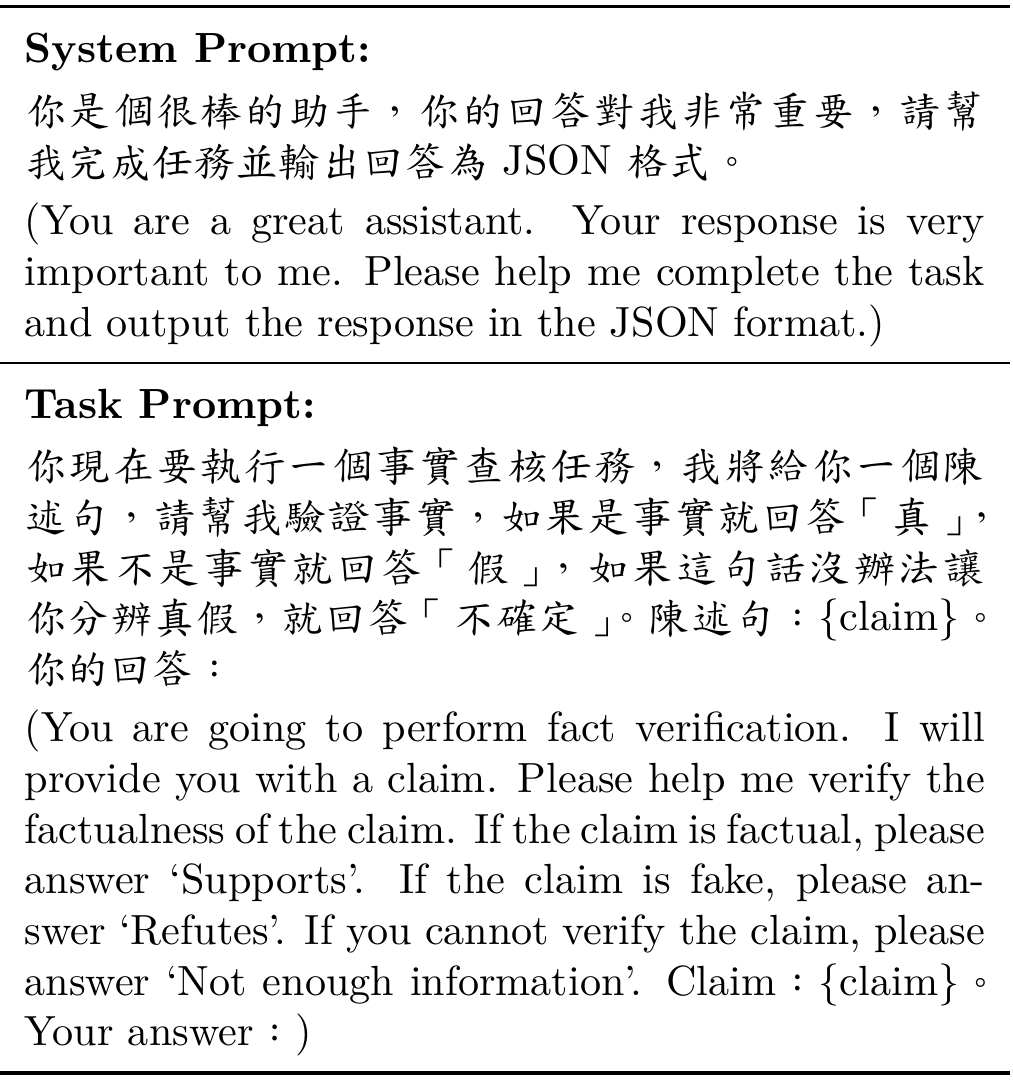}
    \caption{Prompt for the zero-shot and few-shot settings with GPT-3.5 (\textit{gpt-3.5-turbo}). Note that the English part was not included in the experiments.}
    \label{tab:prompt1}
\end{table}
\begin{table}[htbp]
    \centering
    \includegraphics[width=1.0\columnwidth]{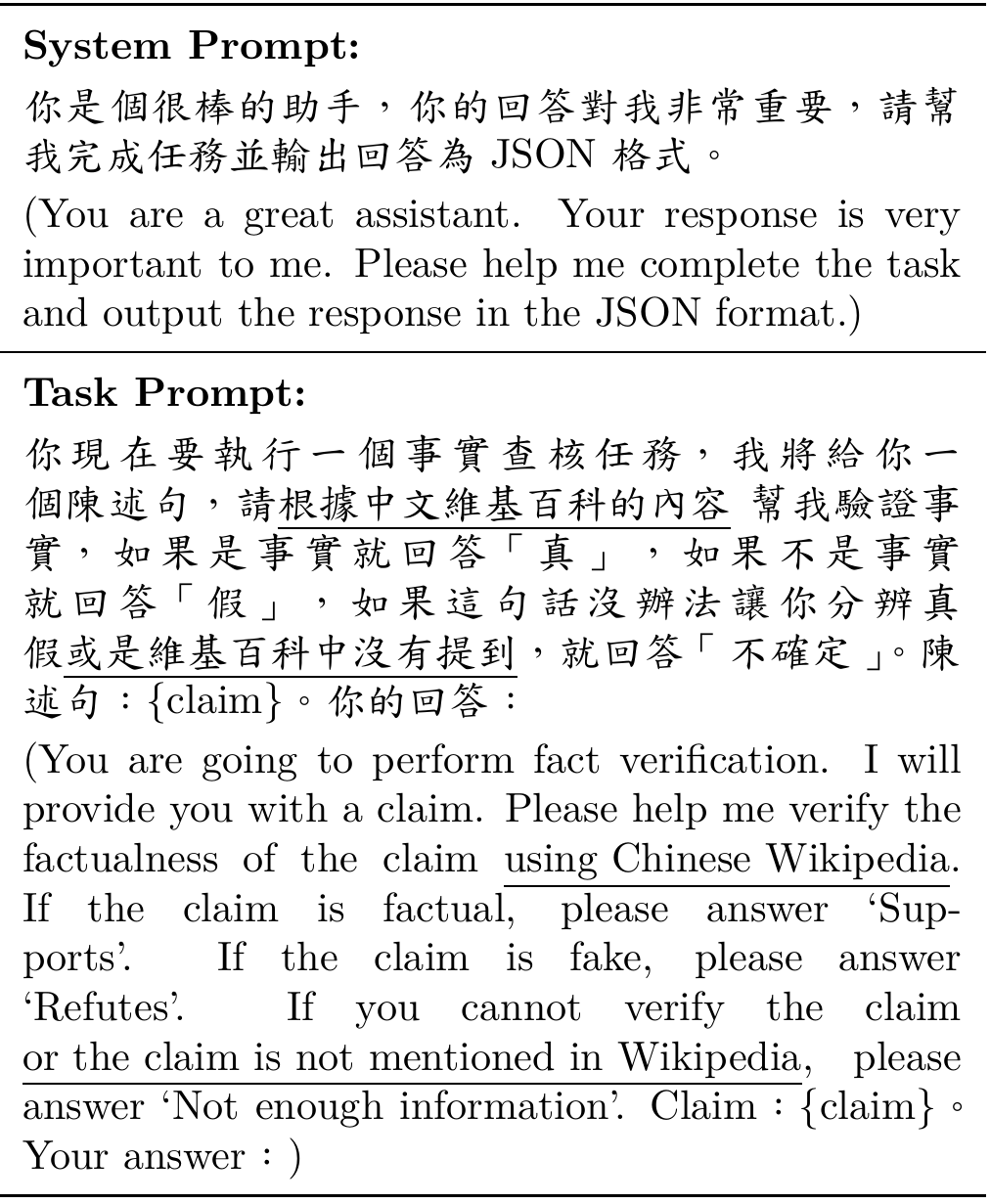}
    \caption{Prompt for the zero-shot and few-shot settings with GPT-4 (\textit{gpt-4-1106-preview}). Note that the English part was not included in the experiments.}
    \label{tab:prompt2}
\end{table}

\subsection{Additional Metrics for the Retrieval Tasks}
In the main content, we provide the recall scores for the document and sentence retrieval results in Table \ref{table:main} and Table \ref{table:gold}.
The corresponding precision and F1 scores are shown in Table \ref{table:ap_retrieval}.
We find that BEVERS has lower precision and F1 scores than our baseline in document retrieval.
The reason is that BEVERS combines the retrieved documents from TF-IDF and the fuzzy string search system \cite{dehaven-scott-2023-bevers}.
The average number of the predicted pages for the claims in the test set is 67.8 in BEVERS's document retrieval results, while we only use the top ten documents for our baseline.
Thus, BEVERS' precision and F1 scores are lower than our baseline scores in document retrieval.

\begin{table}[htbp]
    \centering
    \renewcommand{\arraystretch}{1.2}
    \begin{tabular}{c|lcccc}
        \toprule
        Task      & System        & Recall & Precision & F1    \\
        \midrule\midrule
        Doc       & Our baseline  & 87.65  & 9.79      & 17.62 \\
        Retrieval & BEVERS$^a$    & 92.60  & 2.29      & 4.47  \\
        \midrule
        Sent      & Our baseline  & 76.65  & 25.36     & 38.11 \\
        Retrieval & BEVERS$^a$    & 86.60  & 27.00     & 41.17 \\
        \midrule
        Sent      & Our baseline  & 95.90  & 40.26     & 56.71 \\
        Retrieval & BEVERS$^a$    & 95.20  & 38.98     & 55.31 \\
        (Oracle)  & Stammbach$^b$ & 83.55  & 36.12     & 50.44 \\
        \bottomrule
    \end{tabular}
    \caption{Scores for the document and sentence retrieval tasks. $a$: \citet{dehaven-scott-2023-bevers}. $b$: \citet{stammbach-2021-evidence}.}
    \label{table:ap_retrieval}
\end{table}

\end{document}